# Persona-Based Synthetic Data Generation Using Multi-Stage Conditioning with Large Language Models for Emotion Recognition

Keito Inoshita
*Almondays Co., Ltd*
*Faculty of Business and Commerce*
*Kansai University,*
*Osaka, Japan*
*Email: inosita.2865@gmail.com*

Rushia Harada
*Almondays Co., Ltd*
*School of Economics and Business Administration*
*Yokohama City University,*
*Yokohama, Japan*
*Email: c221219f@yokohama-cu.ac.jp*

***Abstract*—In the field of emotion recognition, the development of high-performance models remains a challenge due to the scarcity of high-quality, diverse emotional datasets. Emotional expressions are inherently subjective, shaped by individual personality traits, socio-cultural backgrounds, and contextual factors, making large-scale, generalizable data collection both ethically and practically difficult. To address this issue, we introduce PersonaGen, a novel framework for generating emotionally rich text using a Large Language Model (LLM) through multi-stage persona-based conditioning. PersonaGen constructs layered virtual personas by combining demographic attributes, socio-cultural backgrounds, and detailed situational contexts, which are then used to guide emotion expression generation. We conduct comprehensive evaluations of the generated synthetic data, assessing semantic diversity through clustering and distributional metrics, human-likeness via LLM-based quality scoring, realism through comparison with real-world emotion corpora, and practical utility in downstream emotion classification tasks. Experimental results show that PersonaGen significantly outperforms baseline methods in generating diverse, coherent, and discriminative emotion expressions, demonstrating its potential as a robust alternative for augmenting or replacing real-world emotional datasets.**

***Keywords—Persona-Based Generation, Synthetic Data, Data Augmentation, Emotion Recognition, Large Language Model***

## I. Introduction

In modern society, a wide range of human activities, such as behaviors, verbal expressions, and purchasing history, are continuously recorded as digital traces. The application of Artificial Intelligence (AI) technologies leveraging these data sources has advanced rapidly. In particular, the field of Natural Language Processing (NLP) has achieved remarkable success in many tasks, largely driven by machine learning using large-scale text corpora [1]. The recent emergence of Large Language Models (LLMs) has further enhanced capabilities in contextual understanding and natural generation, bringing transformative progress to various NLP applications [2].

However, in the area of emotion recognition, which involves the deeper internal states of humans, performance improvements have remained limited. One major obstacle is the severe scarcity of high-quality emotion data, typically comprising emotional labels paired with corresponding text or images. Since emotional expressions are inherently subjective and highly diverse, it is extremely challenging to construct generalizable datasets. In addition, collecting such data is often constrained by ethical and legal considerations. For example, in domains such as workplaces or healthcare environments, practical data acquisition is hindered due to the psychological burden on individuals and privacy concerns. For vulnerable populations, such as children or the elderly, the collection of emotional responses is often considered infeasible under current ethical frameworks.

To address these challenges, there has been growing interest in leveraging generative models to synthesize emotion data. Among these models, an LLM, which is pretrained on extensive text corpora, demonstrates an ability to produce fluent, context-aware emotional expressions when prompted with conditional inputs [3][4]. For instance, by providing specific persona attributes or contextual settings as input prompts, an LLM is capable of generating emotionally rich language that aligns with those conditions. This makes them a promising alternative for generating synthetic data in a manner unconstrained by the ethical and logistical limitations of traditional data collection.

This study introduces a novel framework, PersonaGen, that exploits the capabilities of an LLM to generate diverse and contextually grounded emotion data through multi-stage conditioning. Our approach begins with the construction of base-level persona profiles incorporating attributes such as age, gender, occupation, and personality traits. These are further enriched in a staged manner by integrating socio-cultural dimensions, including educational background, residential region, family structure, religion, belief systems, and income level. Finally, detailed situational information, such as location, activity, interlocutor, communication medium, and linguistic style, is introduced as part of a conditioning prompt, along with target emotion categories, to guide an LLM in generating coherent and natural emotional expressions. This framework enables the creation of high-quality datasets that capture the diversity and subjectivity of emotional language, without the burdens of real-world data acquisition.

To evaluate the effectiveness of the proposed method, we address the following three Research Questions (RQ): RQ1. Can the generated emotional texts exhibit accurate and semantically diverse emotion expressions aligned with the specified emotion categories? RQ2. To what extent are the generated texts perceived as natural and human-like in quality? RQ3. How similar are the generated synthetic data to real-world data, and how effectively can they be applied in downstream emotion classification tasks? Through systematic evaluations of these RQs, this study aims to validate the utility of PersonaGen in both emotion understanding and data augmentation contexts.

The main contributions of this work are summarized as follows:

i) A novel framework, PersonaGen, is introduced, which performs multi-stage conditioning based on layered persona construction to enable the generation of diverse and realistic emotional expressions under rich contextual scenarios.

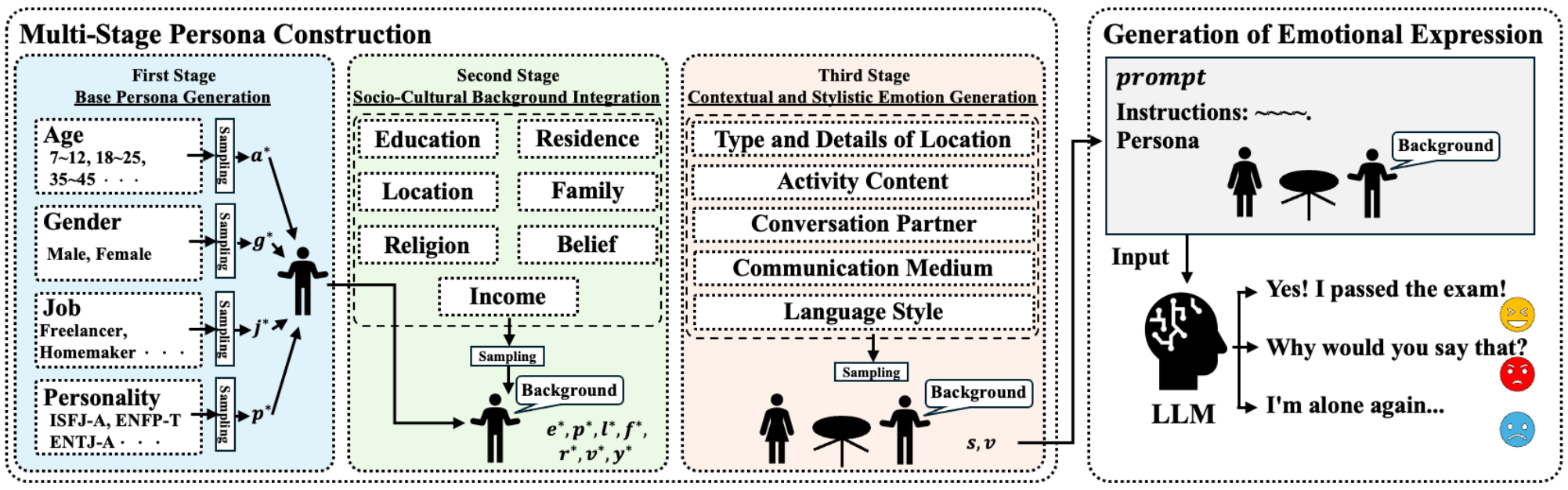

Fig. 1. Overview of the PersonaGen Framework.

ii) The generated synthetic data are quantitatively evaluated from multiple perspectives, such as clustering-based visualization, classification accuracy, semantic similarity, and distributional metrics, which validate the effectiveness and diversity of the generated content.

iii) A comparative analysis with real data is conducted, through which the proposed method is shown to offer practical value in both emotion recognition and data augmentation, especially when real emotional datasets are difficult to obtain.

The rest of this paper is organized as follows. Section II reviews related works. Section III details the proposed method. Section IV presents the experiment setup and evaluation results. Finally, Section V concludes the study and outlines directions for future work.

## II. Related Work

### A. Enhancing Diversity through Persona Control

By incorporating persona-based conditioning into an LLM, it becomes possible to synthesize texts grounded in specific personal backgrounds while enhancing the diversity and naturalness of the resulting emotional expressions. This approach aligns with recent studies focusing on controlling generative behaviors through persona conditioning. Gupta et al. [5] demonstrated that when LLMs are prompted with socially defined personas, reasoning performance tends to decline, and inherent biases become more prominent, which highlights both the power and the potential risk of persona-driven generation. In contrast, Hu and Collier [6] showed that conditioning on attributes such as age and personality significantly improves the reproducibility of subjective response tasks, which supports the effectiveness of attribute-based control in guiding generation. Araujo and Roth [7] conducted a large-scale study involving 162 distinct persona types, revealing that persona influences the content and stylistic features of generated text in a consistent manner, which suggests that multi-attribute personas can serve as a viable source of generation diversity. Similarly, Fröhling et al. [8] found that explicitly specifying personas in prompts leads to diverse yet reproducible outputs in data annotation tasks, which further validates the motivation behind our PersonaGen framework. Kamruzzaman and Kim [9] investigated regional bias in LLM outputs through nationality-based personas and reported that cultural background significantly influences generated text, a finding that aligns with our design choices involving linguistic style and environmental attributes.

On the topic of more structured persona modeling, Li et al. [10] proposed a benchmark for simulating sequential human actions based on detailed profiles of over 1,000 fictional individuals, empirically evaluating the limitations of context-dependent behavior reproduction. Giorgi et al. [11] explored persona-driven generation with both explicit and implicit cues regarding beliefs and life circumstances, providing insights into the capabilities and challenges of subjective content generation. These results collectively suggest that incorporating deeper psychological and personal factors into persona design enhances the semantic richness and diversity of generated output. Finally, Inoshita [12] quantitatively confirmed that an LLM can partially reproduce stylistic features associated with age, based on experiments involving age-specific writing styles.

### B. Generation and Augmentation of Emotion Corpora

The development of reliable emotion recognition models necessitates large-scale, balanced corpora. However, the inherently subjective and multidimensional nature of emotions makes such construction challenging. Consequently, the generation and augmentation of emotion corpora have emerged as active areas of research. Peng and Zhang et al. [13], for instance, proposed a prompting strategy using Chain-of-Thought (CoT) [14] to flexibly manipulate emotional polarity and perspective for data augmentation. Jayawardena and Yapa [15] constructed paraphrased emotional text datasets using GPT, improving both lexical and syntactic diversity. Inoshita [16] synthesized emotion-rich corpora by expanding expressions through rule-based prompting, enhancing semantic variability. These studies all share the goal of utilizing LLMs to generate emotion data, and thus relate closely to the objectives of our study.

Gopali et al. [17] addressed class imbalance in NLP tasks by generating samples for minority classes using GPT, which led to improved classification performance even in emotion-related tasks. Lee and Lee [18] normalized casual expressions from social media using ChatGPT, reporting increased accuracy in emotion recognition, which indicates that LLMs are capable of stylistic adjustments grounded in emotional understanding. Bui et al. [19] generated synthetic emotion corpora by referencing real-world data and combining it with persona-based prompting using an LLM. However, their method relies heavily on existing datasets and risks generating unrealistic personas, which may limit its generalizability and applicability.

## III. Persona-Based Conditioning for Emotional Text Generation

### A. Framework Overview

The overall architecture of PersonaGen, our proposed framework for emotion data generation, is illustrated in Fig. 1. The framework consists of two main components: multi-stage persona construction through conditional generation, and the generation of emotional expressions based on the constructed persona. It comprises four sequential stages, each contributing to the generation of realistic and contextually diverse expressions that reflect real-world variability.

In the first stage, basic personal attributes, such as age, gender, occupation, and personality type, are assigned to construct the foundation of each persona. These attributes are sampled from probability distributions that approximate real-world demographics. To avoid implausible combinations, conditional rules are applied before sampling, and the sampled results are validated afterward. The second stage enriches the persona with socio-cultural background information, including educational attainment, residential region, family structure, religion, belief systems, and economic status. These factors strongly influence emotional tendencies and are essential for constructing personas with diverse and realistic backgrounds. Again, consistency is maintained through predefined constraints and post-generation validation.

The third stage defines specific contextual settings in which the persona exists, including location, activity, interlocutor, media environment, and writing style. This stage plays a critical role in establishing situational context, which facilitates the generation of linguistically coherent and emotionally relevant text. In the fourth stage, all accumulated persona and contextual information are used to prompt the LLM for text generation. The model is instructed to produce language that aligns with both the given emotion category and the constructed scenario, allowing the generated output to reflect the persona's natural emotional response in that context. PersonaGen enables the construction of synthetic data that is both diverse and lifelike, while avoiding the ethical and logistical challenges of real-world data collection. By integrating stepwise persona enrichment with the generative capabilities of high-performance LLMs, this framework offers a novel method for producing diverse, high-quality emotional expressions that are otherwise difficult to obtain through traditional means.

## *B. Base Persona Generation*

In this stage of PersonaGen, each base persona is constructed by conditionally assigning four key attributes: age category $A$, gender $G$, job $J$, and personality type $M$ based on the Myers-Briggs Type Indicator (MBTI) system. The goal is to simulate realistic combinations that reflect actual demographic distributions while supporting diversity in emotion expression generation. Each attribute category is associated with a probability distribution derived from real-world statistics, as shown in Table I. Random sampling is performed according to the probability $P(C)$ of each category. For instance, an age category $A = \{a_1, a_2, \dots, a_n\}$ is sampled as:

$$a^* \sim Categorical\big(P(A)\big) \tag{1}$$

Similarly, the MBTI is sampled from a total of 32 types (16 categories with A/T variants), yielding a personality type $m^*$. The MBTI distribution is based on actual survey data to ensure coverage of diverse personality traits, such as introversion and extraversion.

To prevent implausible attribute combinations, a whitelist-based rule filter is applied in advance. For example, combinations such as "elementary school student" and "lawyer" are excluded. After filtering, gender $g^*$ and occupation $j^*$ are also sampled. The final set of assigned attributes $x = (a^*, g^*, j^*, m^*)$ is then passed to an LLM-based validation process. To detect combinations that cannot be filtered by predefined rules, we employ an LLM to evaluate the naturalness of the constructed persona. The attributes are formatted as a prompt and input into the LLM, which returns one of three labels: "natural," "rare but plausible," or "implausible." Personas labeled as either of the first two categories are retained in the dataset. This process enables the generation of more realistic virtual personas compared to naive random sampling. In addition, LLM-based semantic filtering plays an important role in ensuring contextual coherence among attributes, which enhances the naturalness of the subsequent emotion text generation process.

TABLE I. ATTRIBUTE CATEGORIES AND SAMPLING PROBABILITIES

| Category | Item | Probabilities |
|---|---|---|
| Age | Young Adults | 20.0% |
| | Middle-aged Adults | 25.0% |
| Gender | Males | 49.0% |
| | Females | 49.0% |
| Occupations | Freelancers | 4.8% |
| | Civil Servants | 4.0% |
| MBTI | ISFJ-A | 6.5% |
| | ENFP-T | 2.5% |

## *C. Socio-Cultural Background Integration*

In this stage of PersonaGen, the base persona constructed in the previous step is expanded into a more contextually rich and concrete human model by incorporating socio-cultural background information. This process is essential for capturing the influence of personal values and life circumstances on emotional expressions, thereby enabling the generation of texts rooted in diverse real-world contexts. Seven types of background attributes are introduced in this phase: educational background $E$, prefecture of residence $P$, geographic location $L$, family structure $F$, religion $R$, belief and value system $V$, and income bracket $Y$. These attributes are sampled from probability distributions $P(B)$ that reflect real-world tendencies, with conditional dependencies defined by age category or occupation. For instance, educational attainment is conditioned on the sampled age category $a^*$, such that young adults (e.g., early 20s) are more likely to be assigned "university graduate" or "vocational school graduate," while middle-aged adults are assigned distributions favoring "high school" and "university graduates." This ensures consistency between age and plausible educational background. Similarly, income brackets are designed to reflect typical life stages, such as assigning low-income categories akin to pensions for elderly personas.

TABLE II. BACKGROUND ATTRIBUTES AND DISTRIBUTIONS

| Attribute Item | Category Example | Conditional Distribution Methods |
|---|---|---|
| Educational Background | High school, Graduate school etc. | Adjusted according to the age category |
| Place of Residence | Urban, Rural, Coastal, Island etc. | Based on national population survey |
| Location | Urban area, Suburban, Rural, Inland, Coastal etc. | Based on national population distribution |
| Family Structure | Single, Married, With children, etc. | Based on age and marital status |
| Religion | Buddhism, Christianity, Shinto, No religion, etc. | Based on national religious survey |
| Belief and Value System | Traditional, Progressive, Collectivist, etc. | By education, region, or sampling |
| Income Bracket | 0–1M JPY, 3M–5M JPY, 8M JPY and above, etc. | Based on age range/ student status |

Table II presents representative examples of background attributes and their corresponding sampling strategies. The sampled background attributes $b = (e^*, p^*, l^*, f^*, r^*, v^*, y^*)$ are integrated with the base attributes $x = (a^*, g^*, j^*, m^*)$ to form a unified persona $z = (x, b)$ . To ensure semantic consistency among attributes, we apply an LLM-based validation procedure similar to the one used in the base persona construction stage. Through this structured yet flexible mechanism, a wide variety of enriched personas are generated, each reflecting realistic socio-cultural contexts. This forms a robust foundation for the subsequent stages of PersonaGen, including scenario definition and emotion expression generation, ensuring both contextual richness and data reliability.

### D. Contextual and Stylistic Emotion Generation

In the third stage of PersonaGen, each enriched persona is assigned specific contextual and linguistic settings to facilitate the generation of natural language texts that reflect a designated emotion category. This step aims to simulate the environmental and stylistic conditions under which an emotional expression is realistically produced. Scenario construction is based on five elements, shown in Table III.

TABLE III. ELEMENTS OF SCENARIO DESIGN

| Item Description | Example |
|---|---|
| Type and Details of Location | Café, Factory, etc. |
| Activity Content | SNS posting, Casual chat, etc. |
| Relationship with Conversation Partner | Family, Customer, etc. |
| Communication Medium | Face-to-face, Chat, SNS, etc. |
| Characteristics of Language Style | Polite language, Net slang, etc. |

All elements are sampled from probabilistic distributions to form a concrete scenario $s$ and stylistic configuration $v$. For example, the following configuration may be constructed for a given persona:

- Scenario: "Chatting with a doctor via SNS while waiting in a hospital lobby."
- Style: "Polite tone, youth slang, emotionally expressive, spoken style."

As in the previous stages, these scene-style pairs $(s, v)$ are validated using an LLM to ensure the combinations are plausible. This prevents the inclusion of emotionally expressive texts that are generated under unrealistic or contradictory circumstances.

Once the persona attributes $z$, scene $s$, and style $v$ are finalized, a specified emotion category $e \in E$ is presented to the LLM for text generation. The generation prompt is defined as:

$$prompt = f(z, s, v, e) \quad (2)$$

The output is constrained to a maximum of two short sentences, each of which must clearly reflect the specified emotion.

A notable advantage of this approach is its adaptability to specific application needs. For instance, if one aims to build an emotion corpus focused on SNS behavior, the activity category can be restricted to "SNS posting," and other attributes like location and style can be tuned accordingly to match online communication norms. Conversely, for workplace-oriented studies, scenario filters can emphasize professional locations, activities, interlocutors, and stylistic choices. Thus, this stage of PersonaGen not only enhances the expressiveness of generated emotion texts across diverse personas but also supports high configurability, making it suitable for domain-specific data synthesis and task-oriented dataset construction.

## IV. EXPERIMENT AND ANALYSIS

### A. Experiment Setup

We constructed a synthetic emotion dataset using PersonaGen, a framework that employs multi-stage conditioning based on structured persona construction, and conducted evaluations with respect to the three RQs outlined earlier. All text generation was performed using GPT-4.1-mini by OpenAI, which was selected for its balance between lightweight architecture and strong contextual understanding, making it suitable for generating coherent and diverse emotional expressions. The temperature parameter was set to 1.2 to encourage lexical diversity during generation.

To generate emotion-specific text, a $prompt$ was constructed by combining each persona's attribute profile with a designated emotion category. These prompts were then input into the LLM to produce natural language outputs. The target set of emotion categories, denoted as $E = \{joy, anger, sadness, pleasure, surprise, fear, neutral\}$, was selected based on widely accepted psychological and practical classifications. A total of 3,500 texts (500 per emotion) were generated for evaluation.

### B. Evaluation of Generated Emotional Expressions

To evaluate the semantic diversity and categorical accuracy of the synthetic emotion texts, we first visualized their distribution in embedding space, followed by classification-based analysis using a supervised learning model.

Semantic embeddings were obtained using the all-MiniLM-L6-v2 model [20], after which the vectors were projected to a 2D space using t-SNE. The resulting clusters were color-coded according to emotion categories, as shown in Fig. 2.

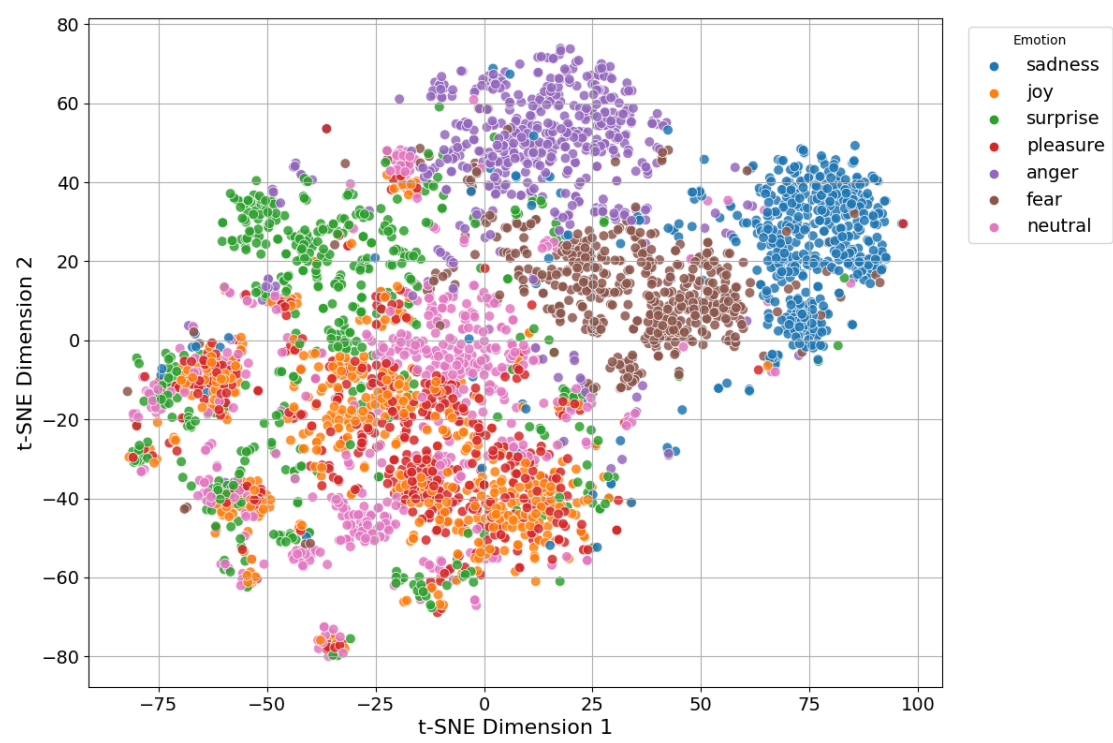

Fig. 2. t-SNE Visualization of Synthetic Emotion Text Embeddings.

The results indicate that categories such as *sadness*, *fear*, and *anger* formed relatively distinct clusters, while *joy* and *pleasure* exhibited partial overlap. This suggests semantic ambiguity between closely related emotional states, especially those sharing similar valence and arousal characteristics. To quantify this variability, we measured three indicators: Mean Cosine Distance (MCD) for lexical spread, Cluster Entropy (CE) for distributional uniformity, and Centroid Distance (CD) for inter-category separation. The results are summarized in Table IV. From the metrics, *surprise*, *anger*, and *neutral* exhibit higher values in MCD and CE, indicating greater variability in expression. In contrast, *joy* and *pleasure* showed more limited diversity, suggesting lexical convergence within those categories.

TABLE IV. EVALUATION ON SEMANTIC DIVERSITY IN SYNTHETIC EMOTION TEXTS

| Emotion | MCD | CE | CD |
|---|---|---|---|
| Sadness | 0.5955 | 1.6028 | |
| Joy | 0.5715 | 1.5716 | |
| Surprise | 0.7361 | 1.5407 | |
| Pleasure | 0.5856 | 1.5821 | 0.4843 |
| Anger | 0.7303 | 1.5948 | |
| Fear | 0.7115 | 1.5564 | |
| Neutral | 0.7476 | 1.5838 | |

Next, we evaluated how well the synthetic texts could be classified into the original emotion categories. A multi-class classification model was trained using LightGBM, taking the same semantic embeddings as input. The dataset was split into 80% training and 20% testing. The classification performance is shown in Table V.

TABLE V. CLASSIFICATION RESULTS OF SYNTHETIC EMOTION TEXTS

| Emotion | Precision | Recall | F1 |
|---|---|---|---|
| Sadness | 0.92 | 0.87 | 0.89 |
| Joy | 0.61 | 0.63 | 0.62 |
| Surprise | 0.81 | 0.88 | 0.85 |
| Pleasure | 0.60 | 0.61 | 0.61 |
| Anger | 0.89 | 0.94 | 0.91 |
| Fear | 0.95 | 0.89 | 0.92 |
| Neutral | 0.87 | 0.80 | 0.83 |
| Total | 0.81 | 0.80 | 0.80 |

Overall, the model achieved strong performance across most categories, with F1 scores exceeding 0.80 except for *joy* and *pleasure*. These results confirm that most synthetic emotion texts are sufficiently distinct to allow accurate classification, though improvements are needed for semantically adjacent emotions.

The confusion matrix in Fig. 3 further illustrates this observation.

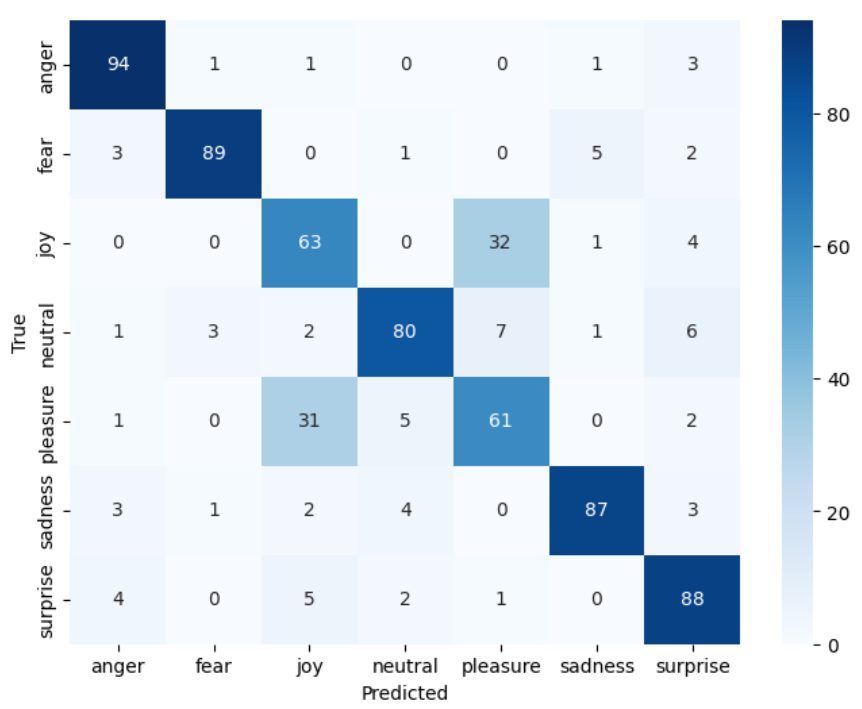

Fig. 3. Confusion Matrix from LightGBM-based Emotion Classification.

A notable degree of confusion was observed between *joy* and *pleasure*, reaffirming their semantic closeness, consistent with the cluster overlap in the t-SNE visualization. Overlap between *neutral* and *pleasure* was also observed, suggesting these categories may share stylistic or content-based similarities.

Taken together, these findings indicate that the proposed method can generate semantically diverse yet structurally distinguishable emotion texts. Particularly, categories like *fear* and *anger* exhibit clear expressive separation, while more affectively ambiguous categories such as *joy*, *pleasure*, and *neutral* may benefit from additional conditioning constraints to enhance separation and clarity.

## C. Evaluation of Human-Likeness in Synthetic Emotion Texts

To assess the human-likeness of the synthetic emotion texts, we utilized an automated evaluation using GPT-4o, which was prompted to assign scores on a five-point scale across four criteria:

- Emotion Match: Whether the text accurately conveys the designated emotional label
- Grammaticality: Whether the sentence structure is grammatically natural, including word order and punctuation
- Lexical Diversity and Appropriateness: Whether the vocabulary is diverse and aligned with the emotional context
- Structure & Logic: Whether the sentence is logically structured and conceptually consistent

The distribution of scores is illustrated in Fig. 4.

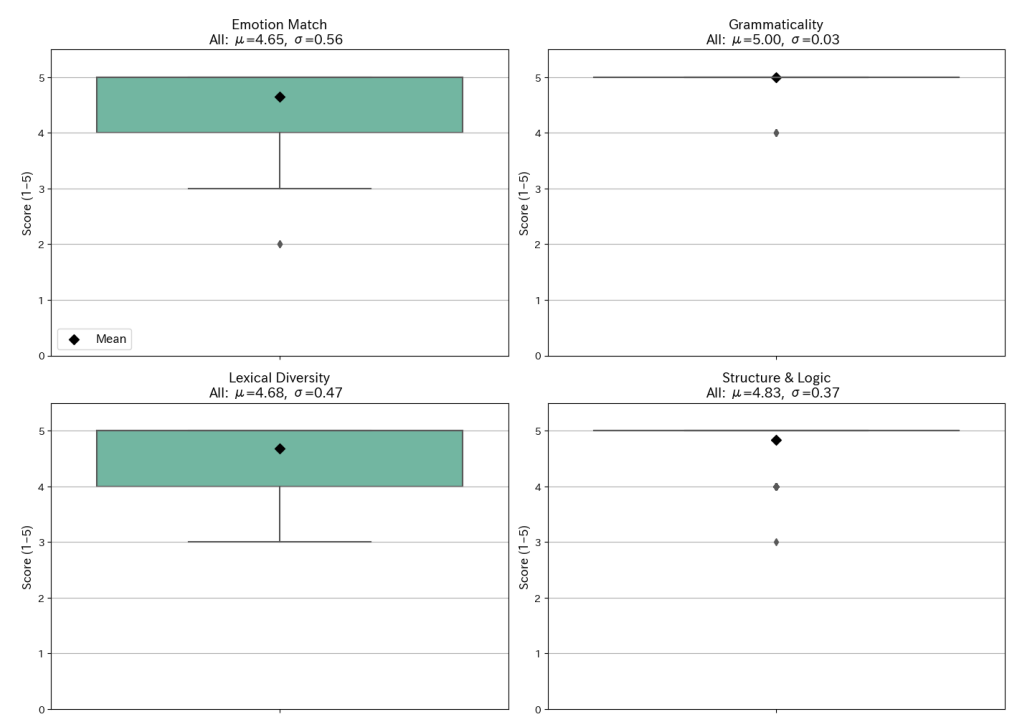

Fig. 4. Evaluation on Human-Likeness Scores.

The results indicate a generally high level of human-likeness across all criteria. Grammaticality achieved a perfect mean score of 5.00, suggesting that nearly all generated sentences were grammatically flawless. Structure & Logic followed closely with a mean of 4.83, indicating strong logical and structural consistency. The scores for Emotion Match and Lexical Diversity also suggest that the texts conveyed the intended emotions using appropriate and varied vocabulary. While minor variability was observed for emotions such as *surprise* and *joy*, the overall results confirm high semantic alignment.

These findings demonstrate that the proposed generation framework produces grammatically correct and semantically rich expressions. This performance can be attributed to the integration of structured PersonaGen conditioning and LLM-based naturalness filtering, which together enhance both contextual coherence and expressive quality.

## D. Evaluation on Real-World Data Comparison

To evaluate the practicality and realism of the synthetic emotion data, we conducted a comparative analysis using real-world data (gold data). For this, we employed a public dataset of emotional tweets available on Kaggle [21], which includes six emotion categories: *joy*, *anger*, *sadness*, *love*, *surprise*, and *fear*. To match this dataset, our synthetic data was generated in tweet format, using explicit scene conditioning, with 500 samples per category (total: 3,000). The following baselines were used for comparison:

- AttrPrompt: A method that generates texts with shuffled attribute combinations to promote diversity.

- Prior Work [22]: A method using randomly generated personas with basic filtering thresholds on LLM outputs.

To evaluate semantic similarity between the synthetic and gold data, we computed the following metrics based on text embeddings:

- Fréchet Inception Distance (FID): Measures distributional similarity in embedding space
- PRD-Fβ: Balances precision and recall of distributions (β=8)
- KL-divergence (KL): Measures asymmetric divergence
- Histogram-Cosine (HC): Quantifies histogram-level similarity

Table VI summarizes the results.

TABLE VI. SEMANTIC SIMILARITY METRICS BETWEEN SYNTHETIC DATA AND GOLD DATA

| Method | FID ↓ | PRD-Fβ ↑ | KL ↓ | HC ↑ |
|---|---|---|---|---|
| AttrPrompt | 0.3877 | 0.2586 | 0.0702 | 0.9824 |
| PriorWork | 0.3967 | 0.2816 | 0.0554 | 0.9858 |
| PersonaGen | 0.4135 | 0.3669 | 0.0489 | 0.9882 |

PersonaGen achieved the best scores in PRD-Fβ, KL, and HC, suggesting it most closely resembled the semantic distribution of the gold dataset. Although AttrPrompt recorded a marginally better FID, our method demonstrated superior contextual consistency and semantic alignment, indicating a more realistic generation framework. This trade-off implies minor room for improvement in statistical homogeneity, while maintaining a strong balance overall.

To assess the utility of the synthetic data in downstream classification tasks, we replaced the training portion of the gold dataset with synthetic data from each method and trained a LightGBM-based six-class emotion classifier. Performance was evaluated on the original gold test set. Table VII presents the classification results.

TABLE VII. EMOTION CLASSIFICATION PERFORMANCE ON GOLD TEST SET USING SYNTHETIC TRAINING DATA

| Method | Accuracy | Precision | Recall | F1 |
|---|---|---|---|---|
| Gold data | 0.568 | 0.569 | 0.568 | 0.567 |
| AttrPrompt | 0.348 | 0.395 | 0.348 | 0.320 |
| PriorWork | 0.326 | 0.379 | 0.326 | 0.288 |
| PersonaGen | 0.359 | 0.400 | 0.359 | 0.321 |

Although none of the synthetic datasets surpassed the performance of gold data, our method consistently outperformed the other baselines across all metrics, suggesting that it retained more discriminative information relevant to emotion classification.

## V. CONCLUSION

This study introduced PersonaGen, a novel framework for emotion text generation based on multi-stage persona conditioning, which enables the synthesis of diverse, context-rich emotional expressions. Comprehensive evaluations demonstrated that the proposed method outperformed baseline approaches in terms of semantic alignment, emotional separability, and human-likeness, validating its quality as a synthetic dataset.

Future work will focus on narrowing the gap with real-world data and improving the practical applicability of synthetic emotion datasets for downstream tasks.

## ACKNOWLEDGMENT

This work was supported by JST SPRING, Grant Number JPMJSP2150.